\documentclass[conference]{IEEEtran}
\IEEEoverridecommandlockouts

\usepackage{amsmath, amssymb, amsthm}
\usepackage{notation}
\usepackage[numbers]{natbib}
\setcitestyle{square}
\usepackage{algorithmic}
\usepackage{algorithm}
\usepackage{mathtools}
\usepackage{notation}
\newtheorem{theorem}{Theorem}
\newtheorem{lemma}{Lemma}
\newtheorem{corollary}{Corollary}

\usepackage[margin=1in]{geometry}
\usepackage{microtype}
\usepackage{graphicx}
\usepackage{subfig}
\usepackage[utf8]{inputenc} 
\usepackage[T1]{fontenc}    
\usepackage{hyperref}       
\usepackage{url}            
\usepackage{booktabs}       
\usepackage{amsfonts}       
\usepackage{nicefrac}       
\usepackage{microtype}      
\usepackage{mathtools}
\usepackage[capitalize]{cleveref}
\crefname{equation}{}{}
\usepackage{algorithmic}
 \usepackage{graphicx}
 \usepackage{placeins}

\usepackage{cite}
\usepackage{amsmath,amssymb,amsfonts}
\usepackage{algorithmic}
\usepackage{graphicx}
\usepackage{textcomp}
\usepackage{xcolor}
\def\BibTeX{{\rm B\kern-.05em{\sc i\kern-.025em b}\kern-.08em
    T\kern-.1667em\lower.7ex\hbox{E}\kern-.125emX}}
\begin{document}

\title{Bounded Expectation of Label Assignment:\\Dataset Annotation by Supervised Splitting with Bias-Reduction Techniques\\
}

\author{\IEEEauthorblockN{Alyssa Herbst}
\IEEEauthorblockA{\textit{Dept. of Computer Science} \\
\textit{Virginia Tech}\\
Blacksburg, Virginia, USA \\
alyssa2@vt.edu}
\and
\IEEEauthorblockN{Bert Huang}
\IEEEauthorblockA{\textit{Dept. of Computer Science} \\
\textit{Tufts University}\\
Medford, Massachusetts, USA \\
bert@cs.tufts.edu
}
}

\maketitle
\begin{abstract}
Annotating large unlabeled datasets can be a major bottleneck for machine learning applications. We introduce a scheme for inferring labels of unlabeled data at a fraction of the cost of labeling the entire dataset. Our scheme, bounded expectation of label assignment (BELA), greedily queries an oracle (or human labeler) and partitions a dataset to find data subsets that have mostly the same label. BELA can then infer labels by majority vote of the known labels in each subset.  BELA determines whether to split or label from a subset by maximizing a lower bound on the expected number of correctly labeled examples. Our approach differs from existing hierarchical labeling schemes by using supervised models to partition the data, therefore avoiding reliance on unsupervised clustering methods that may not accurately group data by label.  We design BELA with strategies to avoid bias that could be introduced through this adaptive partitioning. We evaluate BELA on three datasets and find that it outperforms existing strategies for adaptive labeling.
\end{abstract}
\section{Introduction}
\label{introduction}
A key bottleneck in modern machine learning is the annotation of datasets. While advances in technology have significantly increased the ability of computers to collect large amounts of unlabeled data, supervised learning requires annotation of this data. For example, in classification tasks, this annotation typically requires human experts to provide labels for the true class of each example. The effort and cost of this labeling process is often prohibitive for many applications. In this paper, we introduce a scheme for acquisition of labeled examples that is able to infer high-quality labels with limited labeling budgets. Our approach builds a hierarchy of data subsets via \emph{bounded expectation of label assignment} (\algo). 

Schemes that acquire high-quality labels with lower cost can therefore have tremendous impact on the applicability of machine learning. In settings with a limited budget for annotation, tools are needed to identify which examples would be most informative to label. Such tools need to balance introduction of bias and coverage of the example space. For example, in datasets where different classes have nonuniform proportions, it can be important to ensure coverage of underrepresented classes while avoiding wasteful oversampling of overrepresented classes. 

Random sampling introduces no bias, but it can provide a poor representation of the data space when the budget is low. Random sampling has difficulty finding examples in rare classes or in sparse regions of the input space. On the other hand, active learning approaches \citep{cohn1996active} aim to acquire labels for data most useful for training specific model families. This goal can introduce significant bias \citep{beygelzimer2009importance}. For example, methods such as uncertainty sampling prefer labeling points close to a model's decision boundary, so the distribution of labeled points will be highly dependent on the model family being trained. The acquired labels may not be as useful for training other model families.

Instead, a better approach is to use the structure of the data to determine which examples to label. \citet*{dasgupta2008hierarchical} introduced such an approach called \emph{hierarchical sampling for active learning} (HSAL). They construct a hierarchical clustering and adaptively determine how to prune the clustering. The idea behind HSAL is that if examples sampled from a cluster exhibit high label uniformity, i.e., are mostly the same label, then it can be inferred that the rest of the cluster is likely to have that majority label.
The effectiveness of HSAL thus relies on the quality of the clustering and how well it aligns with the true labels of the classification task. In many settings, a feature-based clustering can have low label uniformity, resulting in negligible gains when using HSAL.

\citet*{dasgupta2008hierarchical} use an unsupervised clustering approach to avoid introducing bias into their data partitioning. However, using some acquired data to guide the partitioning can drastically improve the uniformity of the partitions. Our approach is therefore to design a scheme that can use supervised splitting to partition the data but takes actions to remove the bias induced by such splitting. Supervised splitting allows the algorithm to adapt to information it obtains during labeling, leading to higher quality label inferences and data efficiency.

\algo is the first approach that partitions data guided by a supervised model.  Using supervision is ideal for settings in which clustering does not accurately partition the data by label.  One extreme example of this scenario is when data are uniformly distributed throughout most dimensions of the feature space, in which case even the best clustering methods will fail.  Data are often collected with many measurements irrelevent to the target concept.  Approaches that interactively learn the structure of data can be restricted by the clustering scheme used.  Our approach breaks this restriction by partitioning data with supervised models that can find relevant partitionings.

Our contributions are that we introduce our \algo strategy for actively choosing examples to label. We derive \algo using theoretical analysis to ensure that the actions chosen by \algo greedily improve a lower bound on the expected number of correct labels by the current hierarchy. We define the method to can use either unsupervised or supervised splits and a procedure to ensure that supervised splits do not introduce harmful bias to our labeling scheme. We evaluate \algo on labeling of three datasets, demonstrating that it is able to make good estimates of labels with high data efficiency.

\section{Literature Review}

Active learning is a popular approach for training a learner in the setting where obtaining labeled data is expensive and unlabeled data are abundant.  The learner selects data sequentially to be labeled by an \textit{oracle}, or human labeler. The learner chooses this data by predicting which examples could be most informative.  There are several definitions for the most informative data to label, including least-confidence \citep{li2006confidence}, least-margin \citep{joshi2009multi}, and least-entropy \citep{settles2008analysis}.  Other approaches learn by querying informative and representative examples \citep{huang2010active}.  Unlike traditional active learning in which examples are chosen to be most informative, we seek the most informative \textit{subset} of data, then randomly choose a sample from that subset to label.  We also perform an additional step where we infer labels of unlabeled data if we believe that most data in a subset has the same label, based on the true labels obtained by the oracle. While the typical goal in active learning is to label the data most useful for training a specific model, our goal is to output a labeling of the data that could be useful for training any downstream classifier.

The work of \citet*{dasgupta2008hierarchical} on heirarchical sampling for active learning (HSAL) and that of \citet*{urner2013plal} on probabilistic Lipschitzness active learning (PLAL)
are foundational to ours. The key limitation that has prevented these innovative approaches from having massive impact on the data-hungry state of applied machine learning is that the fixed, unsupervised clusterings that HSAL and PLAL are restricted to often do not fit the label patterns of data. And when they do not fit, the statistical assumptions that are needed for the correctness of their approach prevent any adjustment. Our goal in this work is to build a workaround for this issue, allowing the data partitioning scheme to adapt to the labels it observes, catching data inefficiencies in the label inference scheme. Recent work by \citet*{tosh2018interactive} aims to mitigate the effects of an incorrect clustering by presenting the oracle with a snapshot of a clustering and obtaining a corrected clustering back from the oracle.  However, our approach avoids introducing more complex tasks for the oracle, only asking it to classify individual examples.

The output of \algo is not only a set of inferred labels, but also bounded confidences in those inferences. This type of information makes the labels our method generates amenable to use in weakly supervised learning. Recent methods for weakly supervised learning include approaches that allow annotators to design noisy labeling functions or weak signals and methods that use confidence values to reason about dependencies among the weak signals \citep{ratner2016data,arachie2019adversarial}.

Our method for choosing the most informative subset to sample from relies on concentration bounds on random variables \citep{mcdiarmid1998concentration}. These bounds follow the intuition that if we have seen more labeled samples from a subset of data, we have a more confident estimate of the subset's label distribution.  These bounds guarantee limits on the deviation between empirical measurements of statistics and their true expectations. Our analysis mainly uses Hoeffding's Inequality \citep{hoeffding1994probability}, which is a distribution-free concentration bound---meaning it holds for any underlying distribution.  More specifically, we use a variation of Hoeffding Bounds called the adaptive Hoeffding inequality (AH) proven by \citet*{zhao2016adaptive}:

\begin{lemma}
\label{lemma:adaptive_hoeffding}
Let $X_i$ be zero-mean $1/2$-subgaussian variables. Let $S_\ti = \sum\limits_{i = 1}^\ti  x_i, \ti \geq 1$.  Let $f: \mathbb{N} \rightarrow \mathbb{R}^+$ such that $f(\ti) = \sqrt{a\ti \log(\log_c \ti + 1) + b \ti}$.  Let $c \ge 1$, $a \ge c/2$, $b \ge 0$, and $\zeta$ is the Riemann-$\zeta$ function.  Then,

\begin{equation}
    Pr\big[\exists \ti, S_\ti \geq f(\ti)\big] \leq \zeta(2a/c)\exp(-2b/c)
    \label{eq:adaptive_hoeffding}
\end{equation}
\end{lemma}
The adaptive Hoeffding bound allows a practitioner to decide when to end an experiment as the experiment is occurring, while still maintaining a valid bound.  Traditional Hoeffding bounds require the practitioner to decide on a fixed sample size before samples are obtained.  We derive our bound for the expected number of mislabeled points for a node, based on the labeled data $\labeleddata$ and bounds data $\boundsdata$ for the given node $\node$.

\section{BOUNDED EXPECTATION OF LABEL ASSIGNMENT} 
\label{sec:algorithm}
The algorithm we introduce aims to assign labels to an unlabeled dataset using a limited number of queries to an oracle, or human labeler, which returns the ground-truth label to a single data point.  We provide pseudocode for \algo in \cref{algorithm}.  Due to the limited budget of oracle queries, we cannot obtain labels for the entire dataset and must guess the true labels for some data.  The algorithm searches for \textit{uniform subsets} of data, that is, subsets for which it is believed that most of the ground truth labels are the same.  We refer to data subsets as nodes $\node_i$ of tree $\tree$.  At each step of the algorithm, we will take one of two actions toward the goal of finding uniform nodes: 
\begin{enumerate}
\item \textit{Query} the oracle for the label of a data point, or
\item\textit{Split} a node into child nodes if we are confident that the child nodes are more uniform than the parent.
\end{enumerate}

Our algorithm decides to \textit{split} a node or to \textit{query} the oracle for the label of a data point by estimating which action has a higher potential for finding uniform datasets, and is given by Equation \cref{eq:decision}.  This decision is based on the bounded scores for labeling a new data point \labelcref{eq:boundlabel} or for splitting a dataset \labelcref{eq:boundsplit}.  These scores represent pessimistic estimates of the expected number of correctly labeled points.  When our algorithm terminates, we will assign the majority label of each node to the unlabeled data points within the node.  This label assignment can then be used as input data toward a classification task or as weak supervision.

\subsection{Variable Definitions}

Let $\tree = \{\node_1, ..., \node_{|\tree|} \}$ be the set of leaf nodes of our tree, where each node contains some data points $\node = \{\data_1, ... \data_{|\node|} \}$. Let $\Groundtruth$ be the set of ground truth labels of all examples. At all times during the algorithm, we assume that each data example is assigned the majority label of the leaf node that it belongs to.  This ``majority label'' of a node $\node$ is denoted as $\maj_\node$.  Let $\labeleddata$ be the set of labeled data in a node $\node$ and $\trainingdata$ be the set of ``training data'' used to guide the splitting of a node. $\boundsdata$ is the set of ``bounds data'' used in calculating a pessimistic estimate of correctly labeled data points later in \cref{eq:bound}.  For notation, we use $B_V'$ and $L_V'$ to refer to the bounds data and the labeled data taken from node V'

It is important to note that $\trainingdata$ and $\boundsdata$ are subsets of labeled data within the node and are disjoint. That is, $\trainingdata \subseteq \labeleddata \subseteq \node$, $\boundsdata \subseteq \labeleddata \subseteq \node$ and $\trainingdata \cap \boundsdata = \emptyset$.  We keep the bounds data $\boundsdata$ and training data $\trainingdata$ disjoint so that we reduce bias in our algorithm. We do not evaluate the quality of our splitting technique with the same data that we use to generate the split, because that would cause our algorithm to overstate the quality of the split.

We use $\bound$ to denote the lower bound on the ratio of correctly labeled data points in a node.  There are a couple variations of $\bound$ used in different circumstances, such as $\bound_{\textrm{split}}$ to calculate the speculative bound if we were to split a node and $\bound_{\textrm{split}}$ to calculate the speculative bound if we were to label a data point from a node.  Similarly, we let $\Bound$ denote the lower bound on the ratio of correctly labeled data points in \textit{the entire dataset}, and is given by \cref{eq:bound_all}.

In the calculation of $\bound$ we use $\buffer \in [0, 1]$, an ``error buffer'' from the expected frequency. The probability that the true frequency of $\maj$ exceeds more than $\buffer$ buffer from the observed frequency is $\delta(\buffer)$.  Other variables $a$, $b$, and $c$ used in the calculation of $\bound$ are constants provided by the practitioner.
\section{ALGORITHM DESCRIPTION}

\begin{algorithm}[tb]
\begin{algorithmic}[1]
\caption{\algo}
\label{algorithm}
\REQUIRE Dataset $\Data = \{\data_1, \data_2, ... \data_{N} \}$, $\Groundtruth = \varnothing$, Dataset Tree $\tree$, and oracle labeler $\oracleof(\data)$
\STATE Initialize a root node with all data; $\node_r \gets \{\data_1, \data_2, ... \data_N \}$ 
\WHILE{budget $\budget > 0$}
    \STATE Choose best node $\node$ and action $\action$ corresponding to 
        \begin{equation*}
        \begin{aligned}
            \textrm{Decision}(\node, \action) =& \\
              \argmax_{\node \in \tree, \action \in \{\textrm{split}, \textrm{label}\}} \Big\{& \bound_\action(\node, \boundsdata, \labeleddata) + \\
            &\sum\limits_{\node' \in (\tree - \node)} \bound(\node', \boundsdata_{\node'}, \labeleddata_{\node'}) \Big\}
            \end{aligned}
        \end{equation*}
    \IF{$\action = $ \textrm{label}}
    \STATE \texttt{obtain\_label\_from}$(\node)$
        
    \ELSE
        \STATE $\{\node_1, ... \node_k\} = \splitfcn(\node)$,
        \STATE Remove parent node from tree ; $\tree \gets \tree - \node$
        \STATE Add child nodes to tree; $\tree \gets \tree \cup \{\node_1, ... \node_k\}$
    \ENDIF
\ENDWHILE
\STATE \algorithmicfor ~~ {$\node \in \tree$} \algorithmicdo ~ Set $\groundtruth_{i} = \maj_\node$ for $i$ in $|\node - \labeleddata|$ and add label to return set $\Groundtruth \gets \Groundtruth \cup \{\groundtruth_{i} \}$ \algorithmicend ~ \algorithmicfor
\STATE \textbf{return} $\Groundtruth$
\end{algorithmic}
\end{algorithm}


We maintain a bound, described later in \labelcref{eq:bound_all}, on  the expected number of correctly labeled data points according to the current tree and majority labels for each node. The bound is used to decide for which node to perform an action and which action to perform. At each time step, the node $\node$ and action $\action$ are chosen as follows:
\begin{equation}
\begin{aligned}
    \textrm{Decision}(\node, \action) = \\
    \argmax_{\node \in \tree,\: \action \in \{\textrm{split},\: \textrm{label}\}} & \bound_\action(\node, \boundsdata, \labeleddata) + \hspace{-0.5cm} \sum\limits_{\node' \in (\tree - \node)}  \hspace{-0.5cm} \bound(\node', \boundsdata_{\node'}, \labeleddata_{\node'}).
\label{eq:decision}
\end{aligned}
\end{equation}

We describe the derivation of this estimate formula in \cref{sec:bound}. Since the formula depends on quantities that are either known or can be estimated as the result of split and label actions, we can use the formula to anticipate the resulting estimate after these actions. 

\subsection{Splitting Procedure}
For the split operation, a subroutine $\splitfcn(\trainingdata, \node)$ will divide the examples in $\node$ into a set $\{\node_1, ... \node_k \}$ of child nodes.  The split is guided by the training data $\trainingdata$.  The split operation will also distribute the labeled examples among the child nodes. That is, $\{\node_1, ... \node_k\} = \splitfcn(\trainingdata, \node)$, $\{ \boundsdata_1, ... \boundsdata_k\} = \splitfcn(\trainingdata, \boundsdata)$, and $\{ \labeleddata_1, ... \labeleddata_k \} = \splitfcn(\trainingdata, \labeleddata)$.  For each $\{\node_1, ... \node_k\}$, we empty the isolated training set $\{\trainingdata_1, ... \trainingdata_k\} = \{\varnothing, ... \varnothing \}$.

We further analyze the effect of unsupervised and supervised splitting on our ability to estimate the number of correctly labeled examples in \cref{theoretical_analysis}. 

\subsection{Labeling Procedure}
For the label operation, the subroutine \texttt{obtain\_label\_from}$(\node)$ retrieves the label of a randomly chosen data point from a given node $\node$.  The data points are chosen without replacement such that, for a given split, we will query each data point up to one time.  However, a data point may be queried multiple times as a node is split, as labeled data are passed on to child nodes.   For this reason, we only use our labeling budget if the randomly chosen data point has never been labeled before in a parent node.  The labeled data point is then added to the set of training data $\trainingdata$ with probability $\trainingprob$ or the set of bounds data $\boundsdata$ with probability $(1 - \trainingprob)$. In our experiments we use $\trainingprob = 0.5$.

When performing an oracle query, we anticipate that the oracle will return the leaf node's majority label for the data point queried, thus increasing the count of the majority label and the label count. Therefore, the number of estimated correct labels for the query operation is
\begin{equation}
    \bound_{\text{label}}(\node, \boundsdata, \labeleddata) = \bound(\node, \boundsdata + \data, \labeleddata + \data).
\end{equation}
The full version of this bound is given by \labelcref{eq:boundlabel}.

Should the oracle return a label other than the majority label,  we would expect to make more labeling mistakes, thus decreasing the bound $\bound$.  The majority label may even change if the count of another label is higher than that of the current majority label.  Typically, this results in a temporary dip in $\bound$ until the distribution of the sample approaches the distribution of the entire node.

\section{THEORETICAL ANALYSIS} 
\label{theoretical_analysis}
First, we bound the probability of violating an estimate of the proportion of particular class in a data subset. We obtain the definition of $\buffer$, a proportion of data points that we label incorrectly, and $\delta(\buffer)$, the probability that we obtain an error of $\buffer$.   Later, we derive \cref{thm:bound_node} ourselves, in which we choose the most optimal $\buffer$ and $\delta(\buffer)$ that will yield the highest estimate of correctly labeled data points.

\begin{theorem}
\label{thm:bound_thm}
 Let the labeled data from $\node$ be $\labeleddata$, and the bounds data from $\node$ be $\boundsdata$ such that $\boundsdata \subseteq \labeleddata \subseteq \node$ at time $\ti = |\node - \labeleddata|$.
Let
\begin{equation}
    S_\ti = 
    \frac{|\node - \labeleddata|}{|B|} \sum\limits_{\lab_j \in \boundsdata} \big[ I(\lab_j = \maj) \big] - \sum\limits_{\lab_i \in (\node - \labeleddata)} \big[  I(\lab_i = \maj)\big]    
    \label{eq:sum_vars}
\end{equation}
be the result of a random sampling from node $\node$, and let $f(\ti) = qt$.  Let $c \geq 1$, $a \geq \frac{c}{2}$, and $0 \leq \buffer \leq 1$.  Then,
\begin{equation}
    Pr\big[ \exists \ti, S_\ti \geq f(\ti)\big] \leq \delta(\buffer),
\end{equation}
where
\begin{equation}
    \delta(\buffer) = \zeta (a/c)\exp \Big\{\frac{-2}{c}  \big(\buffer^2 \ti - a\log(\log_c\ti + 1)\big)\Big\}.
    \label{eq:prob_error}
\end{equation}
\end{theorem}

\begin{proof}
Let 
\begin{equation}
    \delta = \zeta(2a/c) \exp(-2b/c) 
    \label{eq:prob_error_short}
\end{equation}
be the probability of encountering a bound violation from the adaptive Hoeffding inequality, \labelcref{eq:adaptive_hoeffding}.  We obtain \labelcref{eq:sum_vars} by centering the count of majority label $\maj$ about the expected frequency of $\maj$ in the bounds data.  We choose $f(\ti) = \buffer\ti$, such that $\buffer$ represents a ``buffer'' frequency from the expected frequency of $\maj$.  That is,
\begin{equation*}
    \begin{aligned}
    \Pr\bigg[ \frac{1}{|B|} 
    \sum\limits_{\lab_j \in \boundsdata} 
    &
    I(\lab_j = \maj) -\\ 
    &\frac{1}{|\node - \labeleddata|} 
    \sum_{\lab_i \in (\node - \labeleddata)}I(\lab_i=\maj) &\geq \buffer  \bigg] \leq \delta .
    \end{aligned}
\end{equation*}
Given that $f(\ti) = q\ti = \sqrt{a\ti \log(\log_c \ti + 1) + b \ti}$, we can solve for $b$:
\begin{equation}
    b = \buffer^2\ti - a\log(\log_c\ti + 1).
    \label{eq:b}
\end{equation}
Substituting \labelcref{eq:b} for $b$ in \labelcref{eq:prob_error_short} gives us our final bound for $\delta$ in \labelcref{eq:prob_error}.
\end{proof}

\begin{lemma}
\label{lemma:no_replacement}
Adaptive Hoeffding Bounds hold when samples are drawn with replacement. \citet*{bardenet2015concentration} state that bounds that hold with replacement will also hold for samples drawn without replacement.  That is,
\begin{equation}
    \mathbb{E}(S_{\ti, nr}) \leq \mathbb{E}(S_{\ti, r}),
\end{equation}
where $S_{\ti, r}$ is $S_\ti$ \labelcref{eq:sum_vars} drawn with replacement and $S_{\ti, nr}$ is $S_\ti$ drawn without replacement.
\end{lemma}

With \cref{thm:bound_thm}, we bound the ratio of a chosen label in an unlabeled dataset of infinite size, of which a finite number of samples have been chosen with replacement. \Cref{lemma:no_replacement} states that the bound obtained from \cref{thm:bound_thm} should hold in the case of samples taken without replacement.  In the next theorem, we apply the bound from \cref{thm:bound_thm} to a dataset of finite size, bounding the proportion of the majority label $\maj$ in the remaining unsampled data of a node $\node$.  

\begin{theorem}
\label{thm:bound_node}
    Let $\node$ be a set of data points and $\maj = \max_{\lab}\sum\limits_{\lab_j \in \boundsdata}I(\lab_j = \lab)$  be the majority label from all labeled bounds data $\boundsdata$. Let $\delta(\buffer)$ be the probability that the proportion of data points with label $\maj$ exceeds $\bound$:
\begin{equation}
    \mathbb{E}\Big[
        \frac{1}{|\node|}\sum\limits_{\lab_i \in \node} I(\lab_i = \maj)
    \Big] \geq \bound(\node, \boundsdata, \labeleddata)
    \label{eq:estimated_correct}
\end{equation}
such that 
\begin{equation}
    \begin{aligned}
    \bound(\node, \boundsdata, \labeleddata) = \max_{\buffer}
        \frac{1}{|V|}
        \bigg\{ &
            |\labeleddata| + 
            |\node - \labeleddata|
            \big(1-\delta(\buffer)\big)
            \times\\
            &
            \Big[
                \frac{1}{|\boundsdata|} \sum\limits_{\lab_i \in \boundsdata} I(\lab_i = \maj)
                - \buffer
            \Big]
        \bigg\}.
    \end{aligned}
    \label{eq:bound}
\end{equation}

We are able to apply adaptive Hoeffding bounds on a finite dataset sampled without replacement using \cref{lemma:no_replacement}.
\end{theorem}
It is important to note that $\delta(\buffer)$ is in fact a parameter to $\bound$.  We choose the $\delta(\buffer)$ that yields the largest value of $\bound$.  This process is described later in 
\cref{sec:bound}.

Using the probabilistic lower bound on the proportion of correct labels for a single node, we can bound the total proportion of correct labels taken over an entire dataset.
\begin{corollary}
The lower bound of the expected proportion of correct labels over the entire dataset is the weighted sum of correct labels over each node:
\begin{equation}
    \mathbb{E}[\# \textrm{correct labels}] \geq \Bound,
\end{equation}
where
\begin{equation}
    \Bound = \frac{1}{|\Data|}
    \sum\limits_{i = 1}^{i < |\tree|} |\node_i|\ \bound(\node_i, \boundsdata_i, \labeleddata_i).
    \label{eq:bound_all}
\end{equation}
\end{corollary}

We derive a speculative bound from \labelcref{eq:bound} for the case in which we choose to label a new data point from a subset.  When labeling, we add a new data point to the set of ground-truth labels and remove a data point from the set of unknown labels.  We assume that the likelihood of obtaining the majority label $L(x = \maj) = \frac{1}{|\boundsdata|} \sum\limits_{\lab_i \in \boundsdata} I(\lab_i = \maj)$ will not change.
\begin{corollary}
\label{cor:bound_label}
The speculative bound for labeling a new data point in a dataset is given as follows:
\begin{equation}
    \begin{aligned}
    \MoveEqLeft[10] \bound_{\textrm{label}}(\node, \boundsdata, \labeleddata)
    = \\
    \max_{\buffer}
    \frac{1}{|V|}
    \bigg\{
        |\labeleddata| +
        1 + 
        &
        \big(
            |\node - \labeleddata|
            - 1
        \big)
        \big(1-\delta(\buffer)\big)
        \times\\
        &
        \Big[
            \frac{1}{|\boundsdata|} \sum\limits_{\lab_i \in \boundsdata} I(\lab_i = \maj)
            - \buffer
        \Big]
    \bigg\}.
    \end{aligned}
    \label{eq:boundlabel}
\end{equation}
\end{corollary}
We can also derive a speculative bound for the case that we decide to split a dataset into $k$ subsets.
\begin{corollary}
Let $\trainingdata$ be a set of labeled training data such that $\trainingdata \subseteq \labeleddata \subseteq \node$ and $\trainingdata \cap \boundsdata = \emptyset$.  Let $\splitfcn(\trainingdata, \node) = \{\node_1, ... \node_k \}$ be a data splitting function that splits $\node$ into $k$ data subsets guided by $\trainingdata$ such that $\node = \bigcup_{i = 1}^k \node_i$.  Then, the speculative bound for splitting a dataset into $k$ subsets is given by:
\begin{equation}
    \bound_{\textrm{split}}(\node, \boundsdata, \labeleddata) = \frac{1}{|\node|} \sum\limits_{i = 1}^k |\node_i| \ \bound(\node_i, \boundsdata_i, \labeleddata_i),
    \label{eq:boundsplit}
\end{equation}
where each $\{\node_1, ... \node_k\} = \splitfcn(\trainingdata, \node)$, $\{ \boundsdata_1, ... \boundsdata_k\} = \splitfcn(\trainingdata, \boundsdata)$, and $\{ \labeleddata_1, ... \labeleddata_k \} = \splitfcn(\trainingdata, \labeleddata)$.
\end{corollary}

Our procedure is designed to increase a pessimistic lower bound on an expected number of correctly labeled examples. As we create leaf nodes containing data subsets, we also track a set of labeled examples for each leaf that are uniformly and independently selected from the leaf population. This unbiased sample allows us to derive a bound on the expected number of correctly labeled points. To avoid introducing bias into the labeled subset, we ``forget'' a node's labeled points when it splits that node. The following sections discuss the bound we use and the reason this forgetting is necessary.

\subsection{Derivation and Analysis of Lower Bound}
\label{sec:bound}
Our goal is to provide a lower bound on the expected number of correct labels obtained by \algo if we choose the majority label for each set of data.  At any given point in \algo, we perform the operation that is expected to yield the largest lower bound.  This bound should be pessimistic to ensure that the bound is rarely violated, but it should still provide a reasonable estimate on the expected number of correct labels.  Most importantly, it should follow the same trend as the true count of correct labels for the algorithm to intelligently decide to label a data point or split a set of data.  

We obtain $\bound$ by taking the expected value of correctly labeled points on the success and failure of the adaptive Hoeffding (AH) bound.  The AH bound represents the probability that the true proportion of $\maj$ in the unknown labels $|\node - \labeleddata|$ falls significantly below the empirical estimate of $\maj$ in the bounds data $\boundsdata$.  When AH holds, we expect to get at least $\Pr(\textrm{label correct})$ correct labels out of all unknown labels with confidence $(1 - \delta)$.  When AH does not hold, we make the pessimistic assumption that we will not make any correct label assignments.  Whether or not AH holds, we will make $|\labeleddata|$ correct label assignments, because these are the ground-truth labels that we have obtained from the oracle.  The expected value of correct labels is 
\begin{equation}
\begin{aligned}
        \MoveEqLeft \mathbb{E}(\textrm{\% correct labels}) \geq \\
        &\frac{1}{|\node|} \big[|\labeleddata| + |\node - \labeleddata|(1 - \delta)Pr(\textrm{label correct}) \big],
\end{aligned}
\end{equation}
where $\Pr(\textrm{label correct})$ is a proportion that is some buffer size $\buffer$ less than the empirical proportion of $\maj$.  Equation \cref{eq:label_correct} is used to derive \cref{eq:bound} in \cref{thm:bound_node}, to calculate the lower bound on the expected value of correctly labeled data points. Equation \cref{eq:label_correct} provides an estimate of the proportion of data points with the majority label.
\begin{equation}
    \label{eq:label_correct}
    \Pr(\textrm{label correct}) = \frac{1}{|\boundsdata|} \sum\limits_{\lab_i \in \boundsdata} I(\lab_i = \maj) - \buffer.
\end{equation}

The number of known data points in $\node$ is $|\labeleddata|$, and the size of unknown data points is $|\node - \labeleddata|$.  Lastly, the probability of AH failure is given by $\delta(\buffer)$.  It is important to note that AH determines both $\delta(\buffer)$ and $\Pr(\textrm{label correct})$.  Both rely on some buffer size $\buffer$.  A larger buffer size $\buffer$ will result in a larger confidence $(1 - \delta)$, but a smaller $\Pr(\textrm{label correct})$.  In a sense, we can perform a tradeoff of correct labels for the confidence that these labels will be correct.  Since our algorithm can choose whichever $\buffer$ we want to perform this tradeoff, it chooses the $\buffer$ that yields the largest $\mathbb{E}(\textrm{\% correct labels})$.  There is no closed-form solution for the maximum of $\bound$.  However, since it is a univariate, unimodal function, ternary search \citep{cormen2009introduction, knuth1997art} can approximate the maximum to the desired accuracy after logarithmic executions of $\bound$. The bound $\bound$ is unimodal because it is the sum of a function linear in $\buffer$ and a product of a concave quadratic function and a log-concave function. Since such a product is itself log-concave and therefore unimodal, and the linear function has no effect on unimodality, the overall bound is unimodal and amenable to ternary search.

Performing these substitutions and choosing the optimal $\buffer$ gives us our lower bound on the expected proportion of correctly labeled points:
\begin{equation}
\begin{aligned}
    \bound(\node, \boundsdata, \labeleddata) = 
    \max_{\buffer}
    \frac{1}{|\node|}
    \bigg\{&
        |\labeleddata| + 
        |\node - \labeleddata| 
        \big(1-\delta(\buffer)\big) \times\\
        &\Big[
            \frac{1}{|\boundsdata|} \sum\limits_{\lab_i \in \boundsdata} I(\lab_i = \maj)
            - \buffer
        \Big]
    \bigg\}.
    \end{aligned}
\end{equation}

Importantly, since each node's estimate is a lower bound on its expected number of correctly labeled examples, the sum of all nodes' estimates also bounds the expected value of the entire tree's number of correctly labeled examples. This fact follows the linearity of expectation. Thus, when \algo chooses the node-action pair that most increases the estimated bound, it is heuristically choosing to improve the estimate of expected total correct labels.

\subsection{Supervised Splits and Preserving Independence}

Equation \labelcref{eq:bound} uses adaptive Hoeffding, which depends on the fact that the random variables are drawn independently from a fixed distribution. Since the random variable in question is whether an example belongs to the majority class, the observed labels satisfy this requirement if they are randomly sampled from any node $\node$. However, if the random points are sampled from a parent node $\node$, which is later split into child nodes $\{\node_1, ... \node_k\}$, special care is needed to ensure that the independence and uniform sampling probability hold.

When the partitioning $\{\node_1, ... \node_k\} = \splitfcn(\node)$ is determined by an unsupervised split, it is not affected by whether points are labeled or what their labels are. Instead, it can be thought of as a pre-determined partitioning. Therefore, a random sample from a parent node $\node$ that happens to be part of a child node $\node_i$ is also a random sample from $\node_i$.

In contrast, when the partitioning $\{\node_1, ... \node_k\} = \splitfcn(\trainingdata, \node)$ is determined by a supervised model, the labeled data from $\trainingdata$ cannot be used to evaluate the split quality because $\trainingdata$ is no longer randomly sampled from the resulting child nodes. \algo tracks an isolated set of labeled examples for each node used to train the splitting models. Isolating this training set 
preserves the fact that the data in the partitioned subsets is randomly sampled, since it is independent of $\boundsdata$ (the labeled data used to calculate the speculative bound $\bound_{\text{split}}$ from \labelcref{eq:boundsplit}).

Regardless of the method of splitting, the usage of the $\bound_{\text{split}}$ and $\bound_{\text{label}}$ scores to decide \emph{whether} to split introduces a dependency. In practice, this dependency can be slight and may not have serious effects on the bounds. However, since the bound $\bound_{\text{split}}$ may be evaluated many times on a node---and using multiple possible partitionings---it is safer to have \algo reset its training data $\{\trainingdata_1, ... \trainingdata_k\} = \{\varnothing, ... \varnothing \}$ after the split operation. Doing so ensures that the data considered for calculating future bounds is uniformly randomly sampled from the new child nodes. The impact of ``forgetting'' labels on data efficiency is somewhat mitigated by the fact that, if an example sampled for labeling has previously been labeled and forgotten, we can simply look up the known label and use it, with no need to invoke the costly oracle.

\subsection{Algorithmic Complexity}
 The dominating computation is the inference to which subset a data point belongs, as BELA is designed for when the budget is much smaller than the total number of unlabeled points, $(\budget < < N)$. Typically, this is linear in the number of points, so in the absolute worst case, the cost of doing these inferences is $O(bN)$.
 
\subsection{Oracle Queries Return Non-Majority Label}
The danger of the oracle queries not returning the majority label comes when the node is split too early. In practice, the algorithm chooses to label a new data point over splitting a data set when the confidence for a particular label is low, such as if there are 2 or more competing labels. We may also choose to end our algorithm after a set minimum number of data points has been sampled from each node.

\section{EXPERIMENTS}
	
We compare \algo to the two prominent active labeling methods: HSAL \citep{dasgupta2008hierarchical} and PLAL \citep{urner2013plal}. Both HSAL and PLAL use unsupervised splitting of data and a tree that does not adapt to new label queries from the oracle.  We show that while HSAL and PLAL achieve a higher accuracy with their initial splits, \algo is able to achieve a higher accuracy overall.  We also recall that the true accuracy of labeled data points is not available to the practitioner at the time the labels are obtained.  The practitioner must rely solely on the bound of the accuracy to provide guarantees about the label quality. We empirically show that this bound $\Bound$ is rarely violated.  
We apply HSAL, PLAL, and \algo to 
MNIST \citep{deng2012mnist}, Fashion-MNIST \citep{xiao2017fashion}, and a synthetic dataset of isotropic Gaussian clusters. We start each method with the training sets of completely unlabeled examples, and we simulate oracle calls by revealing the true label of the requested examples.  For each method, we have a budget of the full dataset size.  That is, we continue to reveal labels until the entire dataset has been labeled by the oracle.  In this way, we can see how early correct labels are found by each of the algorithms.  Each algorithm is similar in that they each provide a lower bound on the number of correct labeled points.  The true number of correct labels found is also provided in the plots.  This is found by counting the correct labels if we were to assign the majority label to each node of the tree.

In addition to comparing \algo to  PLAL and HSAL, we try variations of our algorithm with different $\splitfcn$ functions.  In comparing our method with HSAL we use support-vector machines (SVM) to partition the dataset.  We also use multi-layered perceptrons (MLP), decision trees (DT), and naive bayes (NB).  

We decided to not include uncertainty sampling and other forms of active learning in our experiments because the end goal of other types of active learning differs from ours. Traditional active learning approaches will train a specific model, whereas active labeling infers labels, which can then be used to train any model the practitioner chooses.

 	\begin{figure*}[tb]
    		\centering
    		\subfloat[Digit MNIST true proportion labeled correct.\label{digit_mnist_true}]{	\includegraphics[
    		width=0.35\linewidth
    		]{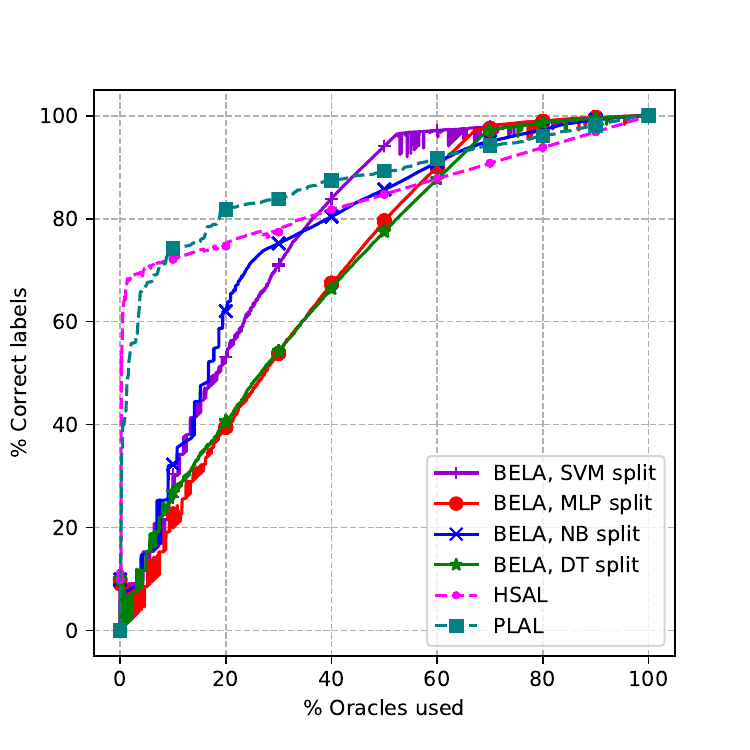}
    		}
    		~~~~~
    		\subfloat[Digit MNIST bound on correct labels.\label{digit_mnist_bound}]{ \includegraphics[width=0.35\linewidth]{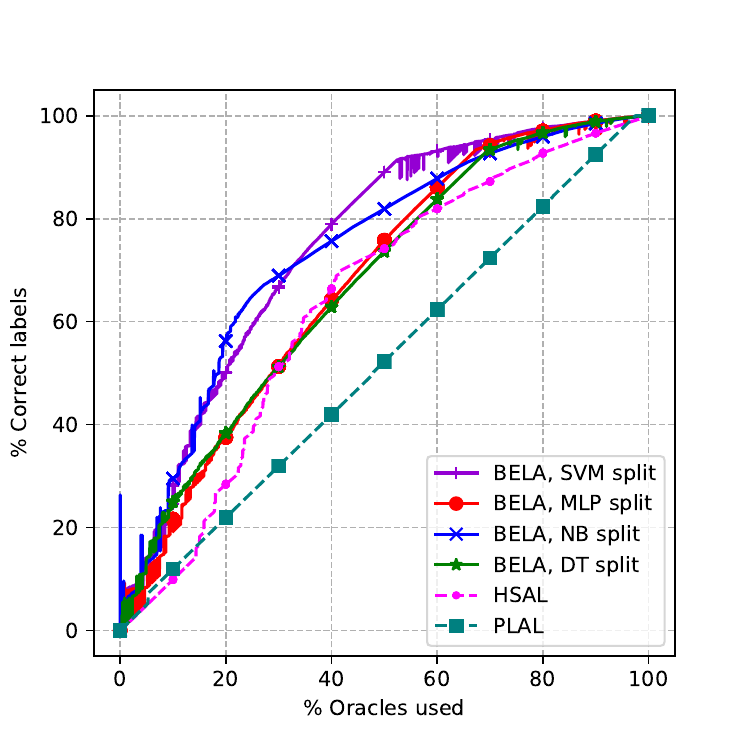}}\\
            \subfloat[Fashion MNIST true proportion correct.\label{fashion_mnist_true}]{	\includegraphics[width=0.35\linewidth]{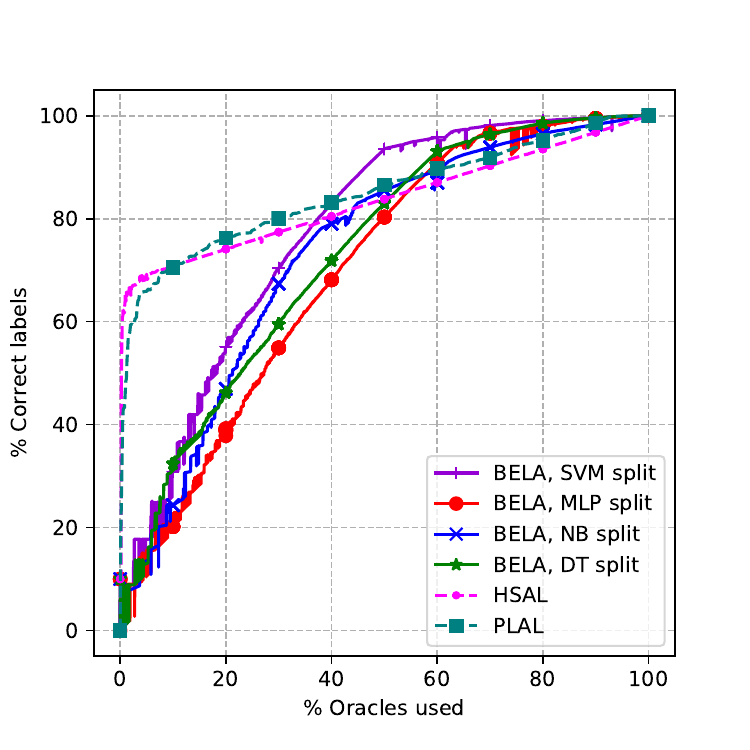}}
            ~~~~~
            \subfloat[Fashion MNIST bound on correct labels\label{fashion_mnist_bound}.]{\includegraphics[width=0.35\linewidth]{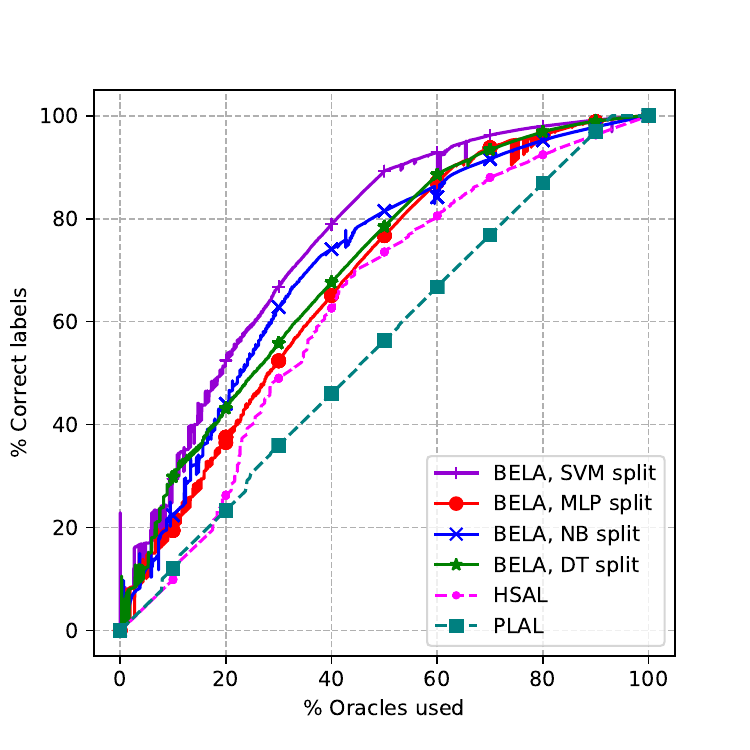}}
		\caption{Digit and Fashion MNIST results. The left plots show the proportion of data that would be correctly labeled if the algorithm stops after various proportions of the oracle budget. This quantity is unknown to the algorithm. The right plots show each method's estimated bound on this proportion, which can be used to decide when to stop, e.g., after reaching a bound estimate of a desired accuracy level.}
		\label{fig:mnistresults}
	\end{figure*}%

\paragraph{Compared Methods}

We describe here how we configured the compared active labeling methods.

We implemented HSAL according to the algorithm given from \citep{dasgupta2008hierarchical}, adapted from the implementation from Kale \citep{kalegithub}.  For the image datasets, we applied PCA and k-means hierarchically to split the dataset into subsets.  While \citet*{dasgupta2008hierarchical} used agglomerative clustering in their experiments, we found that standard implementations of this algorithm are $O(n^3)$ \citep{beeferman2000agglomerative}, which would be intractable for our image datasets of size 60,000.  
Each queried example from the dataset was sampled randomly without replacement.

We implemented PLAL as described by \citet*{urner2013plal}.  We also implemented PLAL with hierarchical PCA and k-means to cluster the image data and hierarchical k-means to cluster the synthetic data.
	
    \paragraph{Datasets}
The MNIST dataset is a set of 60,000 images of handwritten digits.  Each image is 20$\times$20 pixels, is in black and white, and is one of  10 classes of digits $0$--$9$.  
The Fashion MNIST \citep{xiao2017fashion} dataset also contains  60,000 images categorized into 10 classes of different clothing items: T-shirt, trouser, pullover, dress, coat, sandal, shirt, sneaker, bag, and ankle boot.  Each image is slightly larger than digit MNIST at 28x28 pixels and is greyscale.

	\begin{figure*}[tb]
		\centering
    \subfloat[Synthetic data true proportion labeled correct.\label{synthetic_true}]{
    \includegraphics[width=0.375\linewidth]{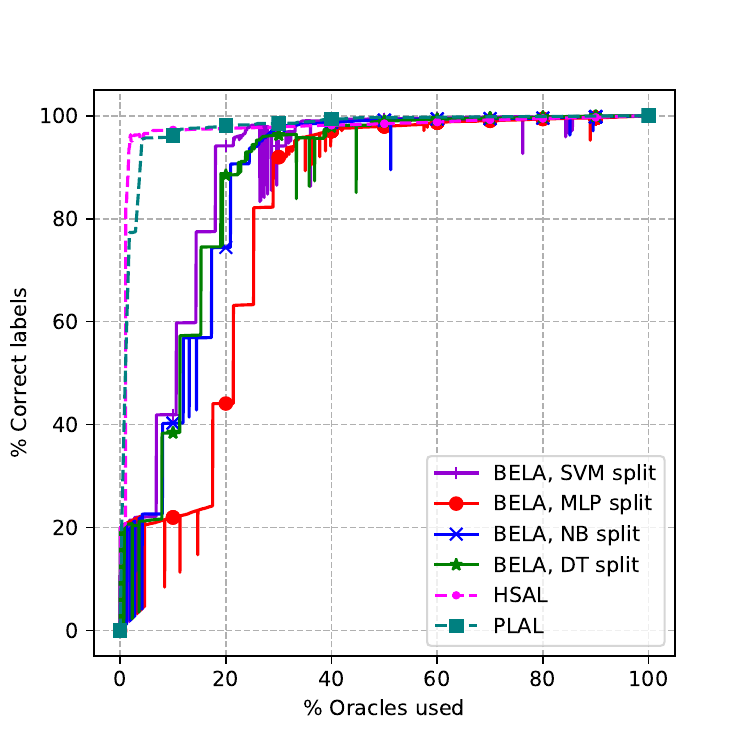}}
    \subfloat[Synthetic data bound on correct labels.\label{synthetic_bound}]{
		\includegraphics[width=0.375\linewidth]{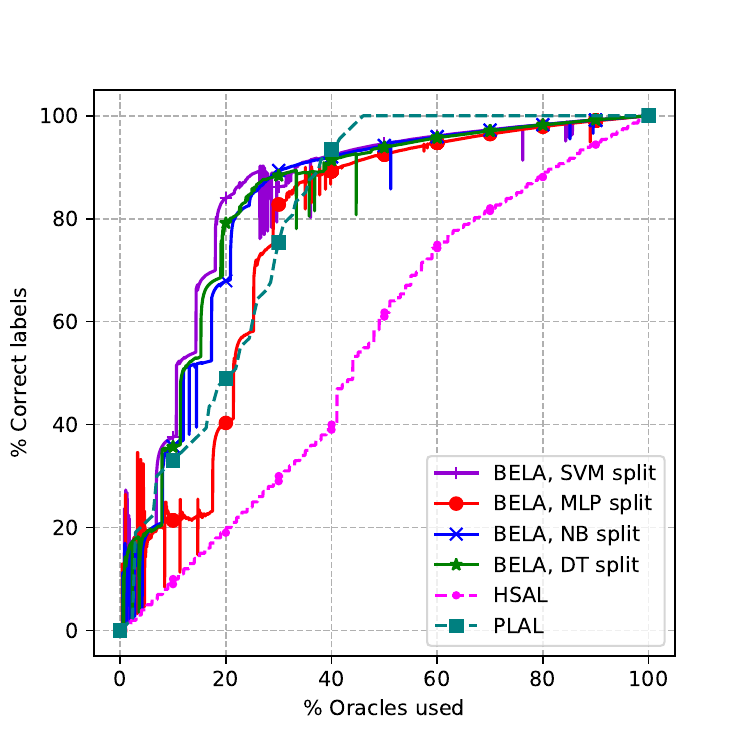}}
	\caption{Synthetic dataset results.}
	\end{figure*}


\subsection{Results} \label{sec:results}
	
	Overall, \algo achieved high accuracy using significantly fewer oracle calls than the compared methods. 
	\algo was also able to provide better guarantees (a higher bound $\bound$) than both PLAL and HSAL at almost all timesteps.  \algo is able to achieve a higher maximum bound than HSAL and PLAL with respect to the bound and the true number of correctly labeled points.  We attribute this advantage to the use of supervision in splits, enabling more accurate partitioning of data subsets.  In an extended version of the paper, we will provide an appendix comparing the lower bound of expected percentage of correct labels and the true percentage of correct labels.  These show that our bound is rarely violated and follows a similar curve as the true percentage of correct labels.
	
	\paragraph{Image Datasets}
	\algo estimates high-accuracy labels using significantly fewer oracle queries than HSAL or PLAL for both image datasets (\cref{fig:mnistresults}). For example, on MNIST, \algo reaches 90\% after querying approximately 45\% of the data, while HSAL requires 70\%, and PLAL requires 60\%. \algo's bound identifies that its accuracy is above 90\% after querying 50\% of data, while HSAL requires 75\% and PLAL requires 100\%. 
	
	Although the HSAL's unsupervised clustering was able to achieve a better split than the \algo's SVM split, it does so only at the beginning of both trials, and the bound for HSAL did not identify that it has made such high-quality a split.  The high-quality initial split can be determined by the higher scores in \cref{digit_mnist_true,fashion_mnist_true} when the percent of oracles calls used is low.  It was necessary for HSAL to obtain more labels from the oracle to confirm the quality of the split.  Comparing the behavior of HSAL and \algo with few oracle calls, HSAL is able to observe the data features without observing the label, whereas \algo must perform a split using only the labeled examples obtained at that time period.  Though HSAL and PLAL use the same $\splitfcn$ function, HSAL still assigns more correct labels than PLAL.  It is important to note that PLAL will not assign labels within a cluster unless \textit{all} the labels obtained within a trial period are the same for a single node.  This trial period constitutes obtaining the labels for $40-150$ data points all at once.  The number of labels obtained depends on the height of the tree, requiring more labels the greater the height of the node.  Most datasets are noisy and will not split data perfectly when partitioned by an unsupervised learning algorithm, meaning that the occurrence of obtaining $40$ data points of the same class is unlikely.  For this reason, PLAL assigns labels almost linearly with respect to the budget used.

	Throughout both trials, the bound for \algo was rarely violated, except for once early in the trial for fashion MNIST.  The bound violations occur when the bound falls below the true proportion of correct labels.
	
	
	\paragraph{Synthetic Dataset}
	HSAL and \algo had similar performance with the synthetic dataset comprised of 2-dimensional Gaussian clusters, however \algo still had a higher estimated bound of correct labels (\cref{synthetic_true} and \cref{synthetic_bound}).  The clusters generated with this dataset were mostly distinct and non-overlapping, meaning it was easily classified with k-means and is an idealized dataset for unsupervised methods like HSAL. Similarly to other datasets, HSAL performed better at guessing the true labels of each data subset near the beginning, but this good guess could not be verified by its bound until nearly all of the budget had been used.

	\subsection{Split Method Evaluation}
	We test \algo using four different $\splitfcn$ functions: linear support-vector machine (SVM), multi-layered perceptron (MLP), naive bayes (NB), and decision tree (decision tree).  Overall, we found SVM and NB to perform best on both image datasets.  All methods performed comparably on the synthetic dataset, with MLP performing slightly worse than the other split methods.  We speculate that SVM and NB performed best because, as low-complexity models, they are less prone to overfitting on a small number of samples compared to nonlinear methods like MLP. When a split is performed, the distribution of the dataset becomes less complex.  Models with low complexity should be able to capture more obvious relationships between dataset features without overfitting on weak relationships between features.  The weaker and less obvious relationships between features may be exploited in future splits of the data.
	
	We also examine the usage of unsupervised and supervised splits.  In almost all experiments, unsupervised splits are able to perform a better initial split on the dataset.  That is, the methods that use unsupervised splitting (HSAL and PLAL) have a higher proportion of true data points labeled correct than do the methods that perform supervised splits.  However, while the proportion of correctly labeled data points may be higher for unsupervised splitting, the lower bound $\bound$ on the correctly labeled data still remains lower than the supervised splitting methods.  The bound does not reflect the initial success of the unsupervised splits.  For both supervised and unsupervised methods, we must test the quality of our splits by querying the oracle for labels.  Later in the experiments, when we have queried higher than 50\% of the dataset labels from the oracle, \algo is able to achieve a higher maximum bound than both PLAL and HSAL (except for in the synthetic dataset).  We believe that this is because supervised splitting is able to perform a more refined split on the dataset that aligns better with the true labels of the data.
	
\section{CONCLUSION} \label{sec:conclusion}
	
We presented a method for finding uniform subsets of an unlabeled dataset by querying labels from an oracle and splitting datasets in a supervised manner.  By finding uniform subsets of data, we may assign labels to be used for future learning tasks without enduring the cost of labeling the full dataset.
Our method uses statistical analysis to reason about when it can stop labeling a data subset with high enough confidence. 

Our framework recursively splits the dataset into subsets that are each assigned a label, seeking sets of data that all contain the same label.
It decides on labeling or splitting actions to increase a confidence-based estimate of the number of correctly labeled points. Our framework uses strategies to allow the use of supervised learning to guide the splitting.

Supervised splitting is preferred in many cases to unsupervised splitting because it is better at finding splits that correspond to the true labels of data, and it is therefore more likely to find uniform clusters.  While supervised splitting induces a bias on the final label assignment, we remove this bias by isolating data to be used to train splits and data to be used to calculate the bounds that score the actions.  In our experiments, our method is able to correctly label significant proportions of datasets while only observing the labels of a small fraction of examples. Moreover, it performs better than similar methods restricted to unsupervised splitting.  With the introduction and evaluation of \algo, we take a key step toward reducing the practical costs of machine learning.

\bibliographystyle{IEEEtranN}
\bibliography{biblio}
\vspace{12pt}

\end{document}